\documentclass[sigconf]{acmart}
\usepackage[capitalise, noabbrev]{cleveref}
\usepackage{algorithmic}
\usepackage{graphicx}
\usepackage{textcomp}
\usepackage{xcolor}
\usepackage{amsmath, bm}
\usepackage{multicol, multirow}
\usepackage{diagbox}
\usepackage{subfig}
\usepackage{balance}
\AtBeginDocument{%
  }

\copyrightyear{2024} 
\acmYear{2024} 
\setcopyright{acmlicensed}\acmConference[ISWC '24]{Proceedings of the 2024 ACM International Symposium on Wearable Computers}{October 5--9, 2024}{Melbourne, VIC, Australia}
\acmBooktitle{Proceedings of the 2024 ACM International Symposium on Wearable Computers (ISWC '24), October 5--9, 2024, Melbourne, VIC, Australia}
\acmDOI{10.1145/3675095.3676609}
\acmISBN{979-8-4007-1059-9/24/10}

\acmSubmissionID{8562}

\begin{document}

\title[$\text{Multi}^3$Net]{Enhancing Inertial Hand based HAR through Joint Representation of Language, Pose and Synthetic IMUs}

\author{Vitor Fortes Rey}
\authornote{Both authors contributed equally to this research.}
\affiliation{%
  \institution{DFKI, RPTU}
  \city{Kaiserslautern}
  \country{Germany}
  }
\email{vitor.fortes_rey@dfki.de}
\orcid{0000-0002-8371-2921}

\author{Lala Shakti Swarup Ray}
\authornotemark[1]
\affiliation{%
  \institution{DFKI}
  \city{Kaiserslautern}
  \country{Germany}}
\email{lala_shakti_swarup.ray@dfki.de}
\orcid{0000-0002-7133-0205}

\author{Qingxin Xia}
\affiliation{%
  \institution{HKUST(GZ)}
  \city{Guangzhou}
  \country{China}}
\email{qingxinxia@hkust-gz.edu.cn}
\orcid{0000-0001-8561-6610}

\author{Kaishun Wu}
\affiliation{%
  \institution{HKUST(GZ)}
  \city{Guangzhou}
  \country{China}}
\email{wuks@hkust-gz.edu.cn}

\author{Paul Lukowicz}
\affiliation{%
  \institution{DFKI, RPTU}
  \city{Kaiserslautern}
  \country{Germany}}
\email{paul.lukowicz@dfki.de}

\renewcommand{\shortauthors}{Vitor Fortes Rey, Lala Shakti Swarup Ray, Qingxin Xia, KaishunWu, \& Paul Lukowicz}

\begin{abstract}
Due to the scarcity of labeled sensor data in HAR, prior research has turned to video data to synthesize Inertial Measurement Units (IMU) data, capitalizing on its rich activity annotations. However, generating IMU data from videos presents challenges for HAR in real-world settings, attributed to the poor quality of synthetic IMU data and its limited efficacy in subtle, fine-grained motions.
In this paper, we propose Multi$^3$Net, our novel multi-modal, multitask, and contrastive-based framework approach to address the issue of limited data. Our pretraining procedure uses videos from online repositories, aiming to learn joint representations of text, pose, and IMU simultaneously. By employing video data and contrastive learning, our method seeks to enhance wearable HAR performance, especially in recognizing subtle activities.
Our experimental findings validate the effectiveness of our approach in improving HAR performance with IMU data. We demonstrate that models trained with synthetic IMU data generated from videos using our method surpass existing approaches in recognizing fine-grained activities.
\end{abstract}


\maketitle

\section{Introduction }
\begin{figure}
    \centering
    \includegraphics[width=1\linewidth]{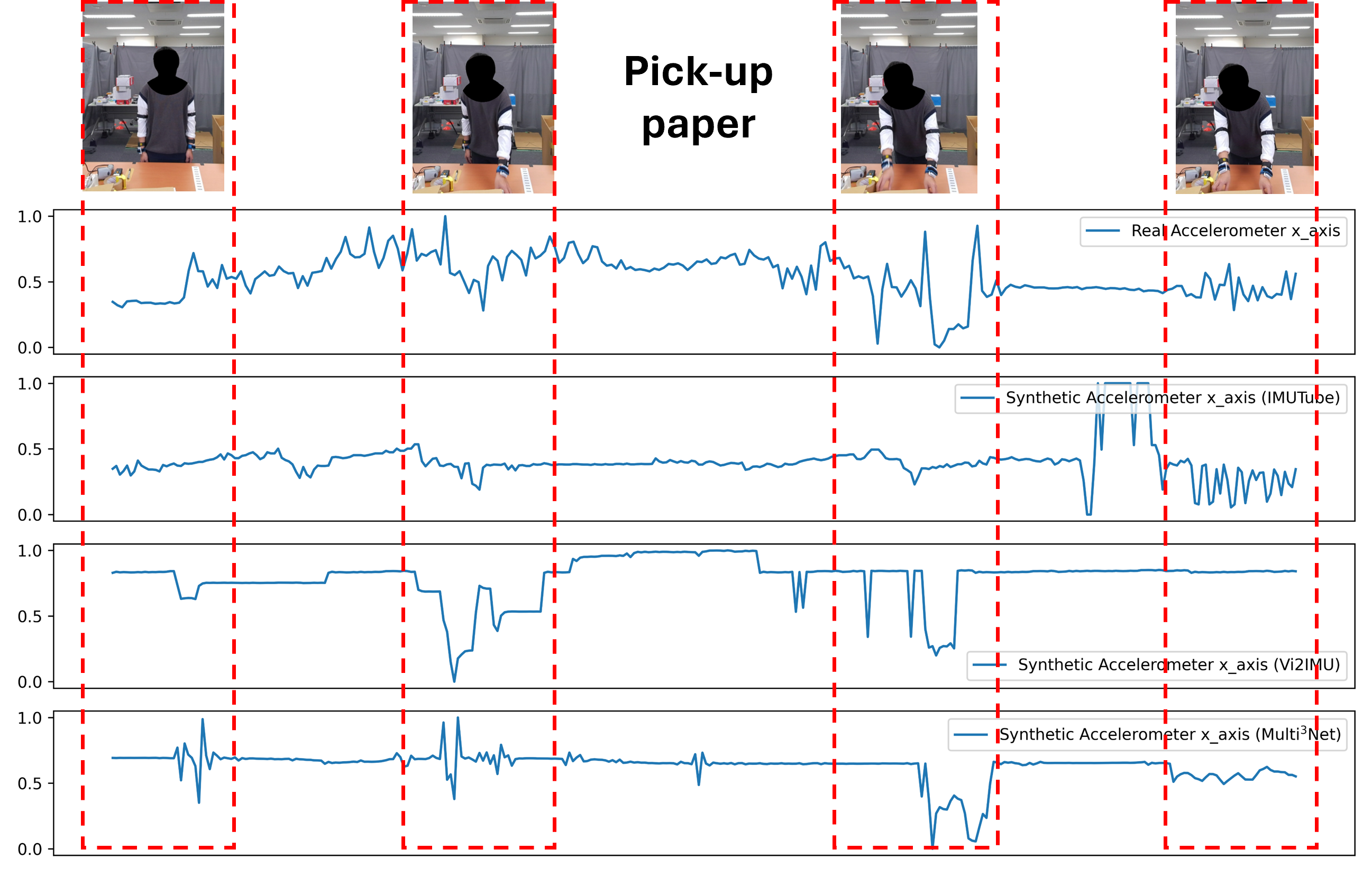}
    \caption{Example of Ground truth IMU data and synthetic IMU data generated by Kinect-based (IMUTube and Vi2IMU) and SMPL-based (Multi$^3$Net) methods. }
    \label{fig:exampleIMU}
\end{figure}


Human Activity Recognition (HAR) using wearable devices has garnered increasing attention in real domains in recent years, such as healthcare \cite{health_sozo}, manufacturing \cite{factory_xia}, and fitness \cite{sport_bo, ray2024text}. 
However, HAR using wearable sensors has benefited much less from recent advances in Deep Learning than fields such as computer vision and natural language
processing. This is, to a large extent, due to the lack of large-scale (as compared to
computer vision) repositories of labeled training data for sensor-based HAR tasks. 
Existing studies have utilized video data to synthesize Inertial Measurement Units (IMU) data to mitigate the labeled data sparsity problem due to its abundance of activity-related annotations, and its effectiveness has been substantiated in \cite{imutube,vi2imu,letimu}, but those methods cannot address fine-grained, subtle movements \cite{imugpt2, utilizeIMU} that are often present in real activities in  complex domains,
such as sticking a label in manufacturing or playing an instrument. 

To utilize the monocular video data and synthesized IMU data for HAR in complex domains one faces two challenges:
\textbf{(1) Low-quality synthetic IMU data}. Existing approaches for generating synthetic IMU data from videos, which require kinetic detection and motion capture, may introduce errors due to lighting, body shapes, and occlusion, thus limiting IMU generation for complex activities. 
Subtle motions, especially on the wrist, can be challenging to simulate from monocular video due to the wrist's degrees of freedom and its relatively small size in pixels.
For instance, as depicted in Figure \ref{fig:exampleIMU}, the synthetic IMU data for the ``pick-up paper'' activity exhibits considerable fluctuations and errors. While simulated data can capture part of the signal characteristics, there is clearly a simulation gap.
\textbf{(2) Directly training on synthetic IMU for HAR results in poor performance for fine-grained activities}. Due to inherent errors mentioned in (1), synthetic IMU data shows heightened sensitivity to activities with smaller amplitudes or greater complexity. Thus models trained with synthetic IMU data may not demonstrate superior performance in HAR.  

While shifts between real and simulated data can harm classification, pose estimation for generating synthetic IMU can provide rich information about the overall body motion and how it is perceived by wearable sensors. In this study, we show how this relationship can be used for  generating better IMU-only representations by leveraging online videos and simulations. By learning to match complex gestures (sign language) to their simulated IMU counterparts, our method can obtain representations with fine-grained activity information that can later be fine-tuned for IMU-only HAR, improving the overall classifier and outperforming other approaches that directly train on simulated data. 

In this paper, we propose \textbf{$\text{Multi}^3$Net}, a multi-modal (text description, Pose, IMU), multi-task (contrastive learning, Pose2IMU generation, IMU reconstruction), and multi-sensor(Left wrist IMU, Right wrist IMU) joint representation framework to enhance the HAR performance in downstream tasks.
Firstly, we employ the Skinned Multi-Person Linear model (SMPL) \cite{smpl} to capture complex human poses with high fidelity and conduct pose calibration to acquire highly accurate synthetic IMU data. 
Then, we leverage contrastive learning for video descriptions$\leftrightarrow$pose, video descriptions$\leftrightarrow$synthetic IMU, and pose$\leftrightarrow$synthetic IMU to learn joint representations across modalities, facilitating the soft adaptation of the pretrained model to target IMU-based HAR tasks. 
Finally, we fine-tune the pretrained model using a small amount of target IMU data for downstream HAR.
The key contributions of this paper are as follows:

(1) $\text{Multi}^3$Net: A multi-modal, multi-task approach to train IMU-only representations using videos, which can learn useful representations for real data training only on simulated ones.

(2) A novel IMU simulation strategy based on SMPL to get better quality IMU data though utilizing fixed bone lengths.

(3) Through extensive experimentation we demonstrate the effectiveness of the proposed approach over other state-of-the-art IMU simulation pipelines through IMU-based HAR.

\section{Related work}

The main bottleneck in developing IMU-based HAR is the lack of rich datasets. Labeled data collection is a cumbersome process due to the time needed for annotation and the variability in sensor setups (position, sampling rate, etc). Recent works have tried to ameliorate this problem either by data simulation or better representation learning.

\textbf{IMU Simulation}:
Various approaches have used other widely available modalities like text and videos to generate the inertial data using 3D simulation and generative models to generate IMU data and use it along with some real data to train HAR models.
In \textit{Video to IMU models}, virtual IMU data was generated from video data \cite{letimu}, initially extracting 2D poses via OpenPose \cite{openpose}, followed by the application of a regression model to translate sequences of poses into IMU data. However, training the regression model requires pre-collected pairs of pose and IMU data. Young et al. introduced IMUSim \cite{imusim} to directly obtain IMU data from sequences of poses.
Nevertheless, the IMU data produced by the above methods often contain considerable errors compared to actual IMU data, thus limiting the utility of using generated data for complex HAR \cite{utilizeIMU, imugpt2}. Santhalingam et al. \cite{vi2imu} proposed a bi-directional LSTM-based model to calibrate anomalous poses with the assistance of surrounding poses, aiming to reduce the amount of incorrectly generated IMU data. However, it does not entirely prevent errors in the translation from pose to generated IMU data.
Liu et al. \cite{videoliu} proposed a CNN-based model to automatically calibrate errors between generated and real IMU data. However, the performance of this approach is influenced by the characteristics of the training data. 
Similarly many \textit{Text to IMU models}\cite{imugpt, imugpt2} have proposed the utilization of textual descriptions converted into human poses for generating IMU data. Various pretrained pose syntheses models, such as MotionGPT \cite{motiongpt}, T2M-GPT \cite{T2Mgpt}, and MotionDiffuse \cite{motiondiffuse}, have been introduced into synthetic human poses, which are then integrated with IMU generation models to obtain synthesized IMU data. However, obtaining the pose is infeasible if the type of pose has never been encountered during the training of the pose synthesis models.
Thus, generating IMU data from video has become the primary approach, especially in professional domains.

\textbf{Representation Learning}:
Contrastive learning has garnered attention in recent years due to its efficacy in learning representations from data originating from diverse domains. For instance, CLIP \cite{clip} learns visual and text representations using paired images and text, achieving outstanding generalization performance for downstream tasks.
Moon et al. \cite{imu2clip} proposed a multi-modal contrastive framework and a pretraining approach to align IMU data with text and video data, thereby projecting multi-modal data representations into a joint space. Yang et al. \cite{videoclip} further enhanced the contrastive framework between text and IMU data by introducing a hierarchical temporal transformer to align important representations.
Different from the aforementioned multi-modal contrastive approaches, our method exclusively utilizes video as input and generates both human pose and highly accurate synthesized IMU data as inputs for the contrastive model. This enables the creation of improved representation spaces for video and IMU data.



\section{Approach}

\begin{figure}
 \begin{center}
\includegraphics[width=0.9\linewidth]{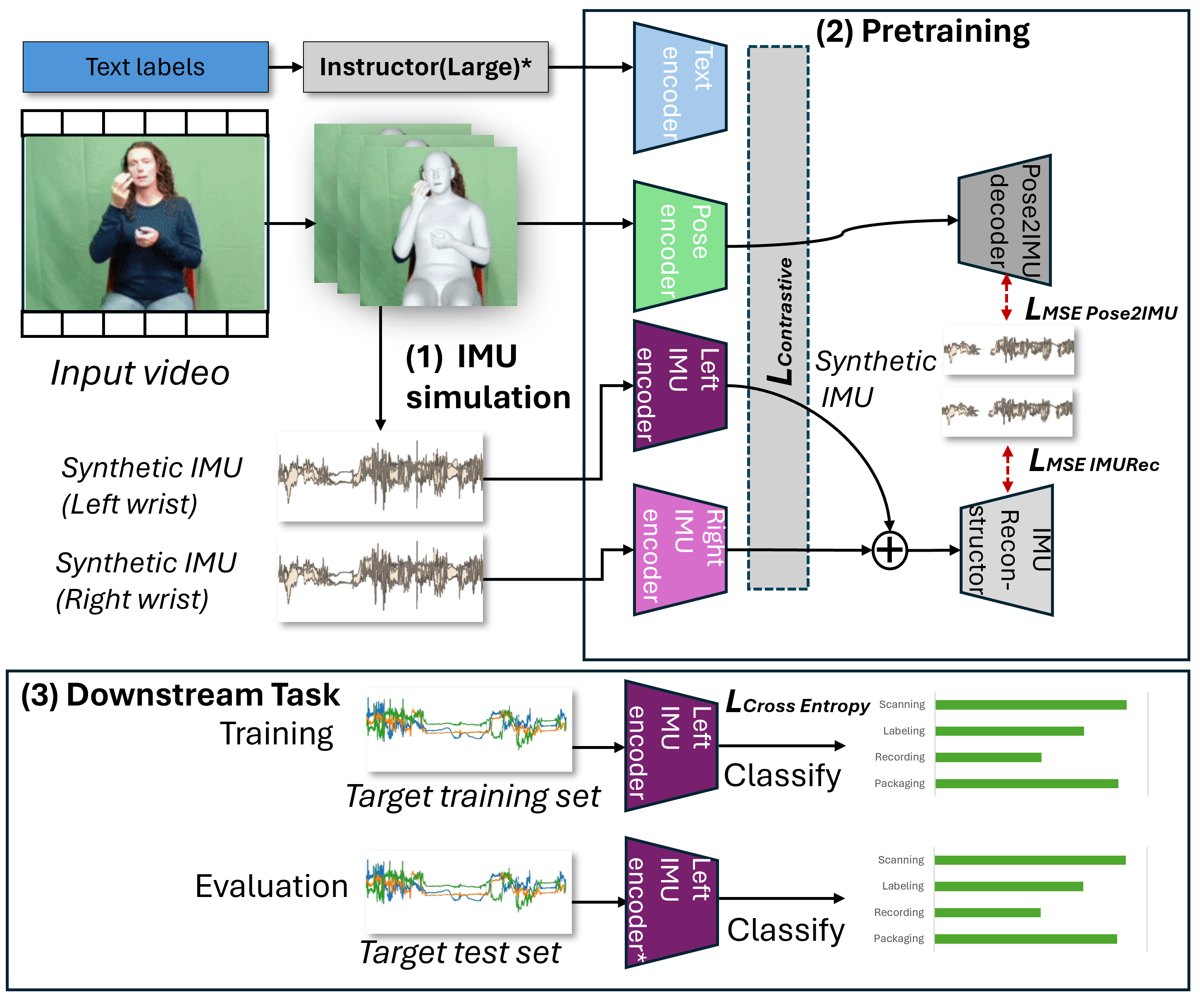}
    \caption{Overview of $\text{Multi}^3$Net architecture showcasing three steps (1) Sensor simulation (2) Multitask pretraining (3) Downstream training and evaluation.}
    \label{fig:overview}
  \end{center}
  \end{figure}
Figure \ref{fig:overview} depicts the overview of the proposed approach, comprising \textcolor{black}{three} primary components: (1) IMU simulation, which generates human poses and high-quality IMU data from video for pretraining; (2) $\text{Multi}^3\text{Net}$, which pretrains the joint text, pose, and IMU representations; and (3) downstream task that involves training to finetune the IMU encoder with real IMU data from the target dataset and evaluating with the target IMU data.



\subsection{IMU Simulation from Videos}

By leveraging human motion capture files, we can precisely calculate both linear acceleration and angular velocity of objects in motion. By tracking the positions and orientations of markers attached to the body over time, we can generate highly accurate data regarding their movement.
The goal of this section is to generate high-quality IMU data from video frames using IMUSim \cite{imusim}, a tool extensively utilized in synthetic IMU generation (e.g., applications such as IMUTube \cite{imutube}, IMUGPT \cite{imugpt,imugpt2}, and etc).
By leveraging human motion capture files calculated from video frames, we can calculate both linear acceleration and angular velocity of objects in motion. 
Tracking the positions and orientations of markers attached to the human body over time enables us to generate IMU data regarding the movement.
To generate synthetic IMU, our approach draws partial inspiration from the Orient3IMU model available in IMUSim removing the noise parameters. We begin with MoCap Motions formatted in SMPL, leveraging Blender to convert the underlying skeleton into an identical shape approximation of specific measurements like height 1.7 meters of average human height for all Mocaps. Subsequently, we relocate the skeleton's origin (Center of Feet) to the (0,0,0) position and orientation.
Transforming linear acceleration to local coordinates considering Gravity can be done using:
\begin{equation}
\mathbf{a}_{\text{local}}(t) = \mathbf{R}_{\text{local}}(t) \cdot \left( \frac{d^2\mathbf{r}_{\text{global}}(t)}{dt^2} - \mathbf{g}_{\text{global}} \right)
\end{equation}
where: $\mathbf{a}_{\text{local}}(t)$ is the linear acceleration of the rigid body in the local coordinate system $\mathbf{R}_{\text{local}}(t)$ is the rotation matrix describing the orientation of the local coordinate system relative to the global coordinates.
$\mathbf{a}_{\text{global}}(t)$ is the linear acceleration of the rigid body in the global coordinate system
$\mathbf{r}_{\text{global}}(t) = [x(t), y(t), z(t)]^T$ is the position vector of the rigid body in global coordinates $\mathbf{g}_{\text{global}}$ is the acceleration due to gravity in the global coordinate system expressed as $\mathbf{g}_{\text{global}} = [0, -9.8, 0]^T$ where $g$ is the magnitude of acceleration due to gravity.
Transforming linear acceleration to local coordinates considering Gravity:

Similarly, after calculating global angular velocity from orientation, we transform it to local coordinates:
\begin{equation}
\boldsymbol{\omega}_{\text{local}}(t) = \mathbf{R}_{\text{local}}(t)^T \cdot \boldsymbol{\omega}_{\text{global}}(t)
\end{equation}
where: $\boldsymbol{\omega}_{\text{local}}(t)$ is the angular velocity of the rigid body in the local coordinate system.

The primary motivation behind adopting our own model over IMUSim lies in the absence of IMU calibration signals necessary for generating good IMU data in IMUSim. Without these signals, the generated data tends to be significantly inferior and out of range of the original signal. Utilizing our model also affords us greater uniformity and control over our dataset. We can ensure consistent starting positions and orientations, as well as uniform human size across all Mocap files, simplifying the neural network's task of establishing correlations within this dataset. 
\textcolor{black}{
Additionally, employing SMPL bodies for pose generation offers advantages to Kinematic 3D pose estimations, as the bone lengths of SMPL bodies remain constant, and the model provides 3D angles rather than identical positions. Furthermore, Kinematic 3D pose estimations require the use of Inverse Kinematics afterward to convert the 3D pose into a Motion Capture file, potentially leading to the loss of information during this process as discussed in the Vi2IMU paper.
}

\subsection{$\text{Multi}^3$Net for Pretraining}
After getting the pose and IMU data generated from video data, we then pretrain joint representations of text, pose, and IMU data via $\text{Multi}^3$Net, which consists of 3 tasks (1) multi-modal contrastive learning, (2) Pose2IMU regression, and (3) IMU reconstruction.

\paragraph{Multi-modal Contrastive learning}
As illustrated in Figure \ref{fig:overview} (2), the pretraining model comprises three encoders, each mapping text, pose, and IMU data to a respective latent space.
Regarding the \textbf{Text encoder}, the input consists of the embedding of the text description of the corresponding video, derived from the output of the last hidden layer of a large pretrained model Instructor (Large) \cite{su2022one}. The output of the Text encoder is denoted as $e_t$. The encoder architecture is based on ResNet architecture with three residual blocks each containing a 1D CNN layer, followed by a batch normalization layer, and a Residual layer.
In contrast to IMU2CLIP, where the text encoder is frozen to facilitate modality transitivity, in our approach, the text encoder is trainable during pretraining to acquire joint representations for multi-modality data.
Similar to the Text encoder, the \textbf{Pose encoder} takes the SMPL pose parameters of the body with (22, 3) tensor except for the two hand parameters, and both left and right hand as Mano parameters (30, 3) tensors to generate the output embedding of $e_{p}$.
The pose encoder is based on the spatial-temporal transformer architecture of PoseFormer \cite{poseformer} where each module is passed to a spatial attention block followed by a temporal attention block to generate intermediate embedding. 
For the \textbf{IMU encoder}, to facilitate adaptable processing across diverse scenarios, we utilize identical multi-headed attention blocks with positional embedding for data collected from both the left and right wrists. The input to the IMU encoder consists of synthetic data segments for each wrist, and the output comprises embedding for the left and right wrists, denoted as $e_{sl}$ and $e_{sr}$, respectively.
Although both encoders share identical architecture the learnable weights are different.

\paragraph{Pose2IMU Regression}
\label{pose2imu}
The Pose2IMU regression block consists of a Pose encoder and a Pose2IMU decoder, which has a CNN architecture with three ConvTranspose and Unpooling layers along with Batch normalization and Dropout blocks.
Since activities in real scenarios typically involve fine-grained motions, the Pose2IMU decoder is designed to guarantee that the input pose encoder encompasses the required features by reconstructing the IMU data from the encoder.
For the Pose2IMU decoder, the input is $e_{p}$, and the output is the predicted result of the corresponding synthetic IMU data, denoted as $X_{p}$. The decoder architecture is based on PSN from PresSim \cite{ray2023pressim}.


\paragraph{IMU Reconstruction}
Similar to the Pose2IMU regression block, the IMU reconstruction block comprises two IMU encoders and an IMU reconstructor. This IMU reconstructor features an identical CNN architecture to that of Pose2IMU but that takes the concatenation of $[e_{sl}, e_{sr}]$ and uses one linear layer to map it back from (256, 2) to 256 vectors to reconstruct one frame input IMU instance. The predicted IMU data is denoted as $X_{s}$.

\paragraph{Loss function}
To acquire joint representations for text, pose, and IMU from the encoders, we propose using instance discrimination by minimizing the InfoNCE loss \cite{infonce} for each pair of encoders, which encourages similar representations for positive pairs closer and pushes representations of negative pairs apart, leading to meaningful feature representations, the InfoNCE loss is defined as follows:
\begin{equation}
 \text{InfoNCE}(q, k) = -\frac{1}{N} \sum_{i=1}^{N} \log \left( \frac{e^{s(q_i, k_i)/\tau}}{e^{s(q_i, k_i)/\tau} + \sum_{j=1,j\neq i}^{N} e^{s(q_i, k_j)/\tau}} \right)
\end{equation}
where
\( N \) is the batch size,
\( q_i \) and \( k_i \) are the representations of the \( i \)-th data sample under two different augmentations,
\( s(q_i, k_i) \) is the cosine similarity score between \( q_i \) and \( k_i \), normalized by the temperature \(\tau\) of 0.07. 
%
The cosine similarity score \( s(q_i, k_i) \) is computed as:
\begin{equation}
 s(q_i, k_i) = \frac{q_i \cdot k_i}{\lVert q_i \rVert \cdot \lVert k_i \rVert}   
\end{equation}
where \( \cdot \) denotes the dot product and \( \lVert \cdot \rVert \) denotes the L2 norm.


Thus, the overall InfoNCE loss of the pairs of encoders is formulated as follows:
\begin{equation}
    \begin{aligned}
        \textbf{L}_{Contrastive} = \text{InfoNCE}(e_t, e_p) + \text{InfoNCE}(e_t, e_{sl})\\
        + \text{InfoNCE}(e_t, e_{sr}) + \text{InfoNCE}(e_{p}, e_{sl}) \\
        + \text{InfoNCE}(e_p, e_{sr}) + \text{InfoNCE}(e_{sl}, e_{sr})
    \end{aligned}
\end{equation}

To ensure that the pose and IMU encoders preserve the necessary features to represent activities, MSE loss is applied for Pose2IMU regression and IMU reconstruction. Taking $X_v$ as the synthetic IMU data, the MSE loss is defined as follows:
\begin{equation}
\begin{aligned}
    \textbf{L}_{MSE} 
    &= \textbf{L}_{MSE \ Pose2IMU} + \textbf{L}_{MSE \ IMURec} \\
    &= \frac{1}{N} \sum_{j=1}^{N} \frac{1}{l} \sum_{i=t}^{t+l} (X^i_v - X^i_p)^2 + (X^i_v - X^i_s)^2 
\end{aligned}
\end{equation}
where $X^i_v$ represents the $i$-th synthetic IMU data point, $X^i_p$ and $X^i_s$ denote the prediction values of the pose2IMU and IMU2IMU decoders, respectively. $l$ represents the segment length.

The overall loss for pertaining is $\textbf{L}_{Contrastive} + \textbf{L}_{MSE}$.

\subsection{Downstream Task}
After pretraining the model using video resources containing rich hand movements, we can subsequently finetune the IMU encoder with a small amount of target IMU data. This process enables us to achieve robust HAR performance even with limited data.
The model structure consists of the pretrained IMU encoder and a classifier.
To prevent the bottleneck problem we take the intermediate output $(6, 256)$ of the pretrained encoder instead of the 1D feature $256$ the final output during pretraining.

We used a hybrid decoder where the classifier processes input feature $(6, 256)$ through a sequence of neural network layers, starting with a CNN layer followed by normalization and ReLU activation. The features are then reshaped and passed through two stages of multi-head attention mechanisms, each followed by normalization and activation, to enable the model to focus on different parts of the input. After each attention stage, the features are further transformed linearly and re-normalized. 
Finally, the processed features are reshaped, and the classes are obtained by averaging across a specific dimension.  



Given a window of sensor data $\bm{X}^{[t:t+l]}$ as the input and $\bm{Y}^{[t:t+l]}=[\bm{y}_t, \bm{y}_{t+1}, ..., \bm{y}_{t+l}]$ as the ground truth label, the classifier is trained to output estimates that have minimum errors to $\bm{Y}^{[t:t+l]}$. 
We train the downstream model using the cross-entropy loss using the Adam optimizer. 
The objective function is formulated as follows:
\begin{equation}
    \textbf{L}_{Cross \ Entropy} = - {\textstyle  {\sum_{i=t}^{t+l}} \sum_{c=1}^{C}} \bm{y}_{ic} \text{log}(p_{ic}),
\end{equation}
where $\bm{y}_{ic}$ is a one-hot vector corresponding to the $i$-th prediction of class $c$, and $p_{ic}$ shows the prediction of $x_l^i$ belonging to class $c$.

\section{Evaluation}
\subsection{Dataset and Training Details}
We utilized two types of datasets (1) Large video datasets with rich hand activity representations for pretraining, and (2) Target inertial HAR datasets with wrist IMUs.
To maintain consistency, all video data were resampled to 60 frames per second.

\textit{How2Sign Dataset \cite{how2sign}} contains more than 80 hours of sign language videos and corresponding transcripts, which provide rich information on hand and wrist movements. This dataset is applied only for pretraining.
\textit{GRAB Dataset \cite{GRAB:2020}} contains approximately 4 hours of MoCap of the entire body of subjects grabbing everyday objects. It comprises data from 10 subjects engaging with 51 different everyday objects.
This dataset is applied only for pretraining.
\textit{OpenPack Dataset\cite{openpack}} contains acceleration data from both the left and right wrists of 5 workers were collected using an Empatica E4 wristband with a sampling rate of 30Hz while they performed a packaging task comprising 11 classes of activities. The workers' activities were also recorded on video to serve as ground truth.
\textit{MM-Fit Dataset\cite{mmfit}} comprises data from 10 subjects engaging in 10 gym exercises. IMU data from Mobvoi TicWatch Pro devices, sampled at 100Hz, capture detailed movement information from the participants' wrists. Additionally, RGB data captured at 30Hz offers visual context for the performed exercises.


\subsection{Quantitative Results}

\begin{table}
\footnotesize
  \caption{Macro F1-score for different datasets.}
  \label{tab:openpack_f1}
  \begin{tabular}{lcc}
    \toprule
    Model&Left wrist& Both wrists\\
    \midrule
    \multicolumn{3}{c}{\textbf{OpenPack Dataset}}\\
    \midrule
    DCL (Real) & \textbf{43.25 $\pm$ 0.81} & \textbf{43.09 $\pm$ 0.50} \\
    DCL (Real + Synthetic Vi2IMU) & 32.71 $\pm$ 0.75 & 33.50 $\pm$ 0.68 \\
    DCL (Real + Synthetic IMUTube) &  42.48 $\pm$ 1.56 &  41.34 $\pm$ 1.48 \\
    Base (Real) & 33.79 $\pm$ 0.39 & {42.26 $\pm$ 0.25} \\
    Base (Real + Synthetic Vi2IMU) & {39.19 $\pm$ 0.18}& 38.28 $\pm$ 0.39 \\
    Base (Real + Synthetic IMUTube) & 35.28 $\pm$ 0.74 & 40.29 $\pm$ 0.89 \\
        \hline
    IMU Reconstruction (how2sign:frozen) & 33.21 $\pm$ 0.53 & 41.19 $\pm$ 0.33 \\
    IMU Reconstruction (how2sign:not frozen) & 39.71 $\pm$ 0.24 & 48.71 $\pm$ 0.45 \\
    Contrastive pretrain (how2sign:frozen) & 39.39 $\pm$ 0.37 & 53.65 $\pm$ 0.18 \\
    Contrastive pretrain (how2sign:not frozen) & 45.27 $\pm$ 0.18& 58.24 $\pm$ 0.26 \\
    IMU Reconstruction (GRAB:frozen) & 31.36 $\pm$ 0.13 & 39.14 $\pm$ 0.19 \\
    IMU Reconstruction (GRAB:not frozen) & 37.33 $\pm$ 0.41 & 46.18 $\pm$ 0.20 \\
    Contrastive pretrain (GRAB:frozen) & 40.21 $\pm$ 0.23 & 53.78 $\pm$ 0.53 \\
    Contrastive pretrain (GRAB:not frozen) & 44.11 $\pm$ 0.26 & 57.17 $\pm$ 0.38 \\
    
    $\text{Multi}^3$Net (how2sign:frozen) & 40.41 $\pm$ 0.17 & 54.17 $\pm$ 0.28 \\
    $\text{Multi}^3$Net (how2sign:not frozen) & 47.32 $\pm$ 0.13& 59.83 $\pm$ 0.27 \\
    $\text{Multi}^3$Net (GRAB:frozen) & 41.22 $\pm$ 0.16 & 55.07 $\pm$ 0.34 \\
    $\text{Multi}^3$Net (GRAB:not frozen) & 45.18 $\pm$ 0.28 & 58.28 $\pm$ 0.28 \\
    $\text{Multi}^3$Net (Both: frozen) & 41.88 $\pm$ 0.28 & 56.36 $\pm$ 0.16 \\
    $\text{Multi}^3$Net (Both:not frozen) & \textbf{48.39 $\pm$ 0.18}& \textbf{61.07 $\pm$ 0.39} \\
    \midrule
    \multicolumn{3}{c}{\textbf{MM-Fit Dataset}}\\
    \midrule
    DCL (Real) &  75.48 $\pm$ 2.53 &  75.79 $\pm$ 2.02\\
    DCL (Real + Synthetic Vi2IMU) &  74.61 $\pm$ 1.78 & 73.73 $\pm$ 2.62\\
    DCL (Real + Synthetic IMUTube)  & 75.63 $\pm$ 1.56 & 75.97 $\pm$ 2.35\\
    Base (Real) & 85.18 $\pm$ 0.31 & 88.13 $\pm$ 0.57\\
    Base (Real + Synthetic Vi2IMU) & \textbf{87.85 $\pm$ 0.48}& 86.59 $\pm$ 0.15\\
    Base (Real + Synthetic IMUTube) & 83.37 $\pm$ 0.26 & \textbf{88.86 $\pm$ 0.25}\\
        \hline
    IMU reconstruction (how2sign:frozen) & 75.63 $\pm$ 0.18 & 78.37 $\pm$ 0.41 \\
    IMU reconstruction (how2sign:not frozen) & 82.74 $\pm$ 0.38 & 86.56 $\pm$ 0.22 \\    
    Contrastive pretrain (how2sign:frozen) & 80.66 $\pm$ 0.61 & 84.52 $\pm$ 0.33 \\
    Contrastive pretrain (how2sign:not frozen) & 89.19 $\pm$ 0.74 & 93.45 $\pm$ 0.71 \\
    IMU reconstruction (GRAB:frozen) & 77.19 $\pm$ 0.34 & 82.11 $\pm$ 0.68 \\
    IMU reconstruction (GRAB:not frozen) & 83.53 $\pm$ 0.64 & 87.15 $\pm$ 0.38 \\
    Contrastive pretrain (GRAB:frozen) & 80.47 $\pm$ 0.49 & 86.64 $\pm$ 0.53 \\
    Contrastive pretrain (GRAB:not frozen) & 88.28 $\pm$ 0.61 & 90.37 $\pm$ 0.18 \\
    $\text{Multi}^3$Net (how2sign:frozen) & 80.55 $\pm$ 0.18 & 86.38 $\pm$ 0.82 \\
    $\text{Multi}^3$Net (how2sign:not frozen) & 91.03 $\pm$ 0.13 & \textbf{93.81 $\pm$ 0.29} \\
    $\text{Multi}^3$Net (GRAB:frozen) & 82.33 $\pm$ 0.17 & 87.59 $\pm$ 0.21 \\
    $\text{Multi}^3$Net (GRAB:not frozen) & 89.72 $\pm$ 0.17 & 91.98 $\pm$ 0.20 \\
    $\text{Multi}^3$Net (Both:frozen) & 81.37 $\pm$ 0.81 & 86.28 $\pm$ 0.26 \\
    $\text{Multi}^3$Net (Both:not frozen) & \textbf{91.15 $\pm$ 0.26} & \textbf{93.44 $\pm$ 0.07} \\
  \bottomrule
\end{tabular}
\end{table}
\raggedbottom

\begin{figure}
    \centering
\includegraphics[width=1\linewidth]{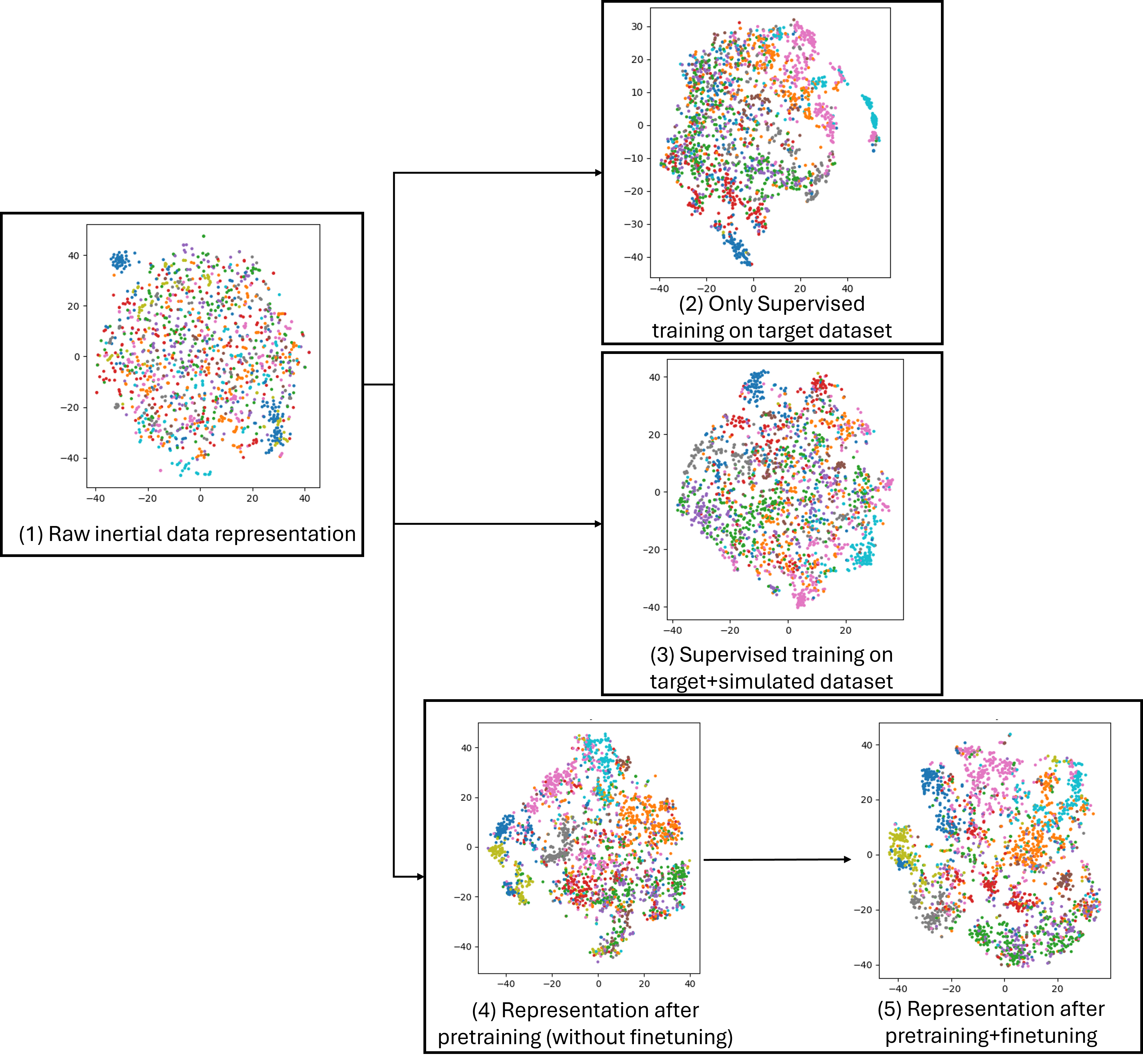}
    \caption{TSNE Latent representations of the proposed approach for OpenPack test set(U0201) where each point depicts a data point in the dataset and each color represents a unique class present in the data.}
    \label{fig:latent_representation}
\end{figure}
In this Section, we compare the proposed approach with synthetic IMU generated by different approaches and directly used for HAR. We conducted leave-one-user-out experiments that were repeated 5 times with different random seeds.


For \textit{DCL (Real)} only real IMU data of the target dataset is used for training, the HAR model is DeepConvLSTM (DCL) \cite{dcl}, which is widely used for IMU-based HAR tasks.
For \textit{DCL (Real+Synthetic Vi2IMU)} both synthetic and real IMU data from the target dataset are used for training. The synthetic IMU is generated by Vi2IMU \cite{vi2imu}, which takes 2D and 3D pose and orientation series as input, and an LSTM-based model is used to predict the virtual IMU. Note that as the orientation is not available in the paper, we utilize OSX \cite{osx} to output the approximate orientations.
For \textit{DCL (Real+Synthetic IMUTube)} both synthetic and real IMU data from the target dataset are used for training. The synthetic IMU is created using IMUTube \cite{imutube}, starting with extracting 2D skeletal poses from videos using AlphaPose \cite{fang2022alphapose}, then mapping the 2D poses to 3D using VideoPose3D \cite{pavllo20193d}, and using IMUSim to generate synthetic IMU data for specific body joints.
Finally the simulated IMU is calibrated using some real IMU data from the target set to have similar range of variability.
For \textit{Base (Real)}, \textit{(Real+Synthetic Vi2IMU)}, and \textit{(Real+Synthetic IMUTube)} we do the downstream training without any pretraining. The IMU encoder weights are initialized randomly. We use real IMU data, real and synthetic IMU data generated by Vi2IMU and IMUTube, to train the model. 
In case of \textit{IMU reconstruction (frozen)} and \textit{(not frozen)} only the IMU reconstruction model is applied for pretraining where "\textit{frozen}" indicates that the learned weights of the IMU encoder are frozen, while
"\textit{not frozen}" means that the IMU encoder is frozen for N epochs until the loss stops decreasing and reaches the patience P for the first time, after which it is unfrozen while the classifier is learning.
In \textit{Contrastive pretrain} only the multi-modal contrastive model is applied for pretraining.
\textit{Multi}$^3$\textit{Net} is the proposed approach that utilizes multi-task pretraining to create a better joint representation for downstream classification.
Tables \ref{tab:openpack_f1} display the HAR performance for the OpenPack and MM-Fit datasets, respectively. Overall the proposed approach outperforms the methods that directly utilize real IMU or real and synthetic IMU data for HAR with one or both wrists. \textbf{Specifically, the proposed method demonstrates an 18.81\% macro F1 improvement for OpenPack with one wrist and 14.6\% for both wrists. The gain in Macro F1 for MM-Fit with one wrist is of 5.96\% while with two wrists it is of 5.31\%.} 
This improvement is significant when compared to synthetic IMU generated approaches, which in some cases decrease the performance. This discrepancy may be attributed to errors in calculating the joint orientation, indicating that the quality of synthetic IMU data significantly impacts HAR performance for complex activities.


\subsection{Discussion and Limitations}


\textbf{Effect of Pretraining}.
Pretraining with simulated data in fact helps in creating better clustering of different latent representations. As we can see in Figure \ref{fig:latent_representation}, the combination of pretraining followed by fine-tuning yields the highest discriminability of latent representations in IMU data. While some clustering is observable in Figures 3 (2) and (4), these representations appear less structured compared to Figure 3 (5).
In contrast, Figure 3 (1) illustrates scattered clusters, suggesting that raw IMU data without pretraining and fine-tuning lacks clear patterns in latent representation.
Even if we are pretraining on a rich dataset that is not of the target it still improves the overall results. 
Also, our method, not frozen, consistently improves on our baseline, which is not always true for other simulation methods.
 Additionally based on the results we can clearly see that single-task pretrainings like Contrastive pretrain and IMU reconstruction are outperformed by multi-task pretraining used in $\text{Multi}^3$Net regardless of the type of pretraining or downstream dataset.
Hence demonstrating how multi-task pretraining is superior to the single-task approach.

\textbf{Effect of Freezing Encoder Weights}
Based on the results presented in the tables, the "not frozen" approach outperforms the "frozen" approach on both datasets. This indicates that optimizing the encoders pretrained from a different data modality (i.e., video) is beneficial for IMU-based HAR tasks.
This is to be expected as the pretraining datasets share non standard activity types and do not really contains both the specific activities present in the target dataset. Given differences regarding IMU calibration and  signal ranges indeed we expected poorer results when the encoder is not learning anything from the target dataset.
Nevertheless $\text{Multi}^3$Net can improve the overall HAR when we allow encoders that acquired some knowledge during the pretraining process to be later fine-tuned with knowledge related to the downstream task, with this approach consistently outperforming other methods.

\textbf{Effect of Real Data Amount}.
\begin{figure}
    \centering
    \includegraphics[width=0.99\linewidth]{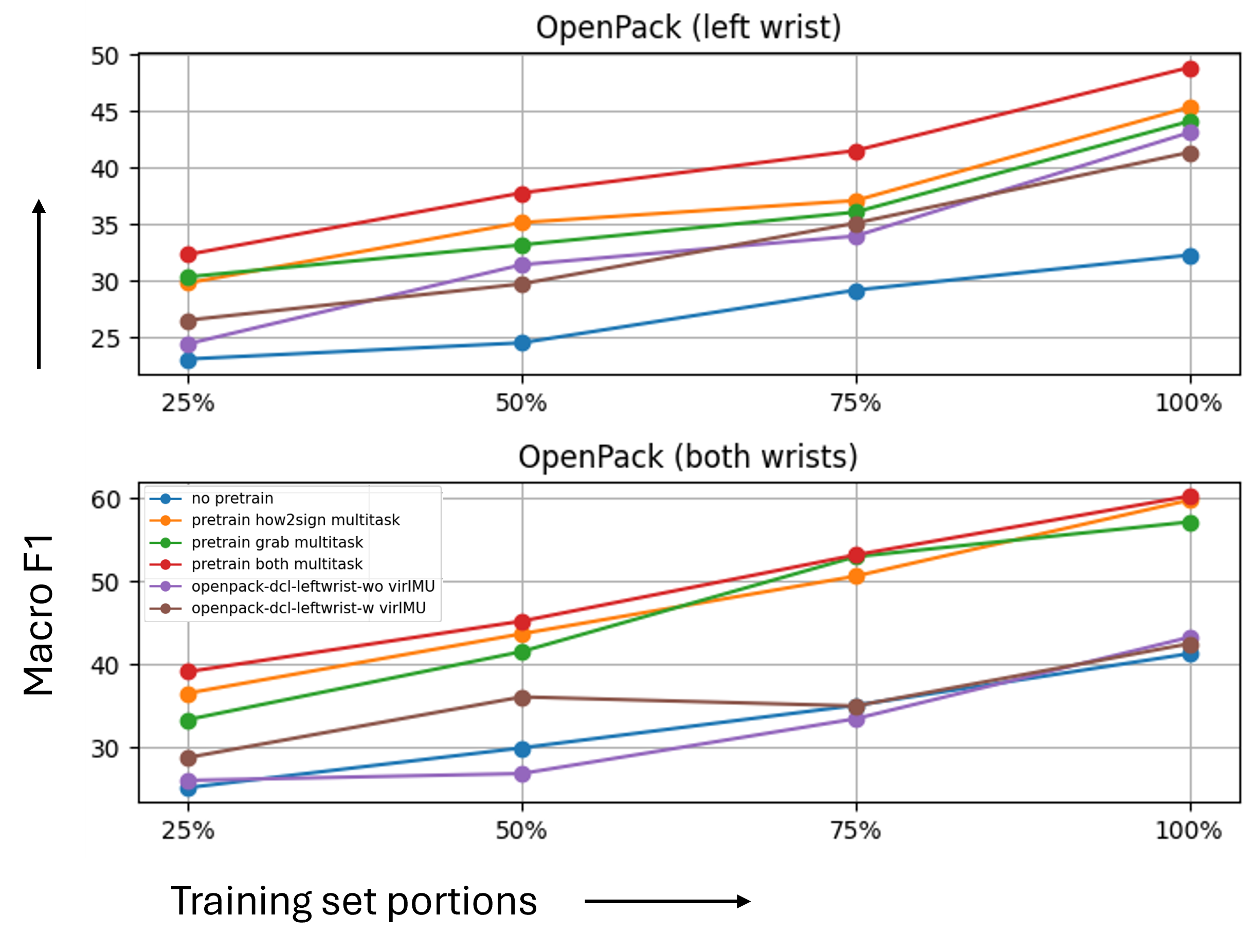}
    \caption{Macro F1-score on different amount of IMU used for downstream task (left wrist: top,both wrists: bottom) using Baseline, DCL (only real data, real+virtual data IMUTube) and pretrained $\text{Multi}^3$Net .}
    \label{fig:amount_realimu}
\end{figure}
We conduct experiments on the OpenPack dataset to investigate how the size of the training set influences the macro F1-score of our approach.
As depicted in Figure \ref{fig:amount_realimu}, our method is label efficient, with similar or better performance than the full training set with 25\% of it. As the amount of real IMU data increases, the macro F1-score of the proposed approach also increases, as opposed to virtual data generation approaches, that are close or worse than their baselines at higher data rates.


\textbf{Limitations}.
As shown in the tables, different datasets for pretraining have an impact on the downstream HAR performance, highlighting the importance of selecting appropriate datasets for pretraining. Furthermore, it is observed that the performance on relatively simple activities does not show a significant increase.
This lack of enhancement can likely be attributed to the fact that the pretraining datasets predominantly consist of non-standard activity types, which may not effectively generalize to more straightforward activities. This discrepancy suggests that the pretraining process might not be adequately capturing the nuances required for these simpler tasks.
Additionally, as illustrated in Figure \ref{fig:exampleIMU}, the simulated IMU data exhibits notable imperfections, whose severity correlates with the accuracy of the estimated pose. This dependence underscores the critical role of pose estimation quality in shaping the fidelity of simulated IMU data. Consequently, the downstream performance of HAR models may be affected by the fidelity of the simulated IMU data, necessitating careful consideration and refinement of pose estimation techniques in conjunction with HAR model development.
We also need further study on the effect of the Base model. Our performance may also be currently limited by the baseline architecture, as it alone does not outperform DCL in the OpenPack dataset. 



\section{Conclusion}

This paper presents $\text{Multi}^3$Net, an extensive framework that leverages multi-modal contrastive learning to enhance IMU-based HAR performance using video data (sign language), with activity classes distinct from the target datasets. The IMU simulation approach is utilized to effectively extract fine-grained activity features from the video, and the learned features enhance downstream HAR tasks, particularly in recognizing complex activities with limited data availability. 
Experimentation conducted for fine-tuning demonstrates that the "not frozen" setting consistently outperforms the others, exhibiting from  5\% to 18\% improvement compared to baseline methods. 
In the future, we plan to investigate the impact of pretraining data selection, the challenge of accurately estimating poses for high-fidelity IMU simulation as well as comparing our method to other multi-modal pretraining approaches such as MESEN\cite{MESEN}.


\begin{acks}
This paper was supported by the BMBF (German Federal Ministry of Education and Research) in the project VidGenSense (01IW21003), the Carl Zeiss Stiftung under the Sustainable Embedded AI project (P2021-02-009) and 
China NSFC Grant U2001207, Guangdong Provincial Key Lab of Integrated Communication (No.2023B1212010007), the Project of DEGP (No.2023KCXTD042).
\end{acks}

\newpage
\bibliographystyle{ACM-Reference-Format}
\balance
\bibliography{reference}
\end{document}